%% file: emnlp2018.tex
\documentclass[11pt,a4paper]{article}
\usepackage[hyperref]{emnlp2018}
\usepackage{times}
\usepackage{latexsym}
\usepackage{amsfonts}
\usepackage{afterpage}
\usepackage{xcolor}
\usepackage{graphicx}
\usepackage{subcaption}
\usepackage{paralist}
\usepackage{url,bm}
\usepackage{amsmath, cancel}
\usepackage{multirow}
\usepackage{url}
\aclfinalcopy 

\newcommand{\rnR}{\mathbb{R}}
\newcommand\norm[1]{\left\lVert#1\right\rVert}

\title{Learning Gender-Neutral Word Embeddings}

\author{Jieyu Zhao  \qquad
  Yichao Zhou    \qquad
  Zeyu Li \qquad
  Wei Wang \qquad
  Kai-Wei Chang\\
  University of California, Los Angeles \\
  \{jyzhao, yz, zyli, weiwang, kwchang\}@cs.ucla.edu
  }

\date{}

\begin{document}
\maketitle
\begin{abstract}
\input{./abstract.tex}
\end{abstract}

\input{./intro.tex}

\input{./related_work.tex}

\input{./methodology.tex}

\input{./experiments.tex}

\input{./conclusion.tex}

\bibliography{emnlp2018}
\bibliographystyle{acl_natbib_nourl}



\end{document}

%% file: abstract.tex
Word embedding models have become a fundamental component in a wide range of Natural Language Processing (NLP) applications. 
However, embeddings trained on human-generated corpora have been demonstrated to inherit strong gender stereotypes that reflect social constructs. 
To address this concern, in this paper, we propose a novel training procedure for learning gender-neutral word embeddings. Our approach aims to preserve gender information in certain dimensions of word vectors while compelling other dimensions to be free of gender influence. Based on the proposed method, we generate a Gender-Neutral variant of GloVe (GN-GloVe). Quantitative and qualitative experiments demonstrate that GN-GloVe successfully isolates gender information without sacrificing the functionality of the embedding model.

%% file: intro.tex
\section{Introduction}{\label{tab:intro}}

Word embedding models have been designed for representing the meaning of words in a vector space. These models have become a fundamental NLP technique and have been widely used in various applications. However, prior studies show that such models learned from human-generated corpora are often prone to exhibit social biases, such as gender stereotypes~\cite{kaiwei,caliskan2017semantics}. 
For example,  the word ``programmer'' is neutral to gender by its definition, but an embedding model trained on a news corpus  associates ``programmer'' closer with ``male'' than  ``female''. 

Such a bias substantially affects downstream applications.  \newcite{zhao2018gender} show that a coreference resolution system is sexist due to the word embedding component used in the system. This concerns the practitioners who use the embedding model to build gender-sensitive applications such as a resume filtering system or a job recommendation system as the automated system may discriminate candidates based on their gender, as reflected by their name.
Besides, biased embeddings may implicitly affect downstream applications used in our daily lives. For example, when searching for ``computer scientist'' using a search engine, as this phrase is closer to male names than female names in the embedding space, a search algorithm using an embedding model in the backbone tends to rank male scientists higher than females', hindering women from being recognized and further exacerbating the gender inequality in the community.

To alleviate gender stereotype in word embeddings, \newcite{kaiwei} propose a post-processing method that projects gender-neutral words to a subspace which is perpendicular to the gender dimension defined by a set of gender-definition words.\footnote{Gender-definition words are the words associated with gender by definition (e,g., mother, waitress); the remainder are gender-neutral words.} However, their approach has two limitations. First, the method is essentially a pipeline approach and requires the gender-neutral words to be identified by a classifier before employing the projection. If the classifier makes a mistake, the error will be propagated and affect the performance of the model. Second, 
their method completely removes gender information from those words which are essential in some domains such as medicine and social science~\cite{back2010gender,mcfadden1992study}.


To overcome these limitations, we propose a learning scheme, Gender-Neutral Global Vectors (GN-GloVe) for training word embedding models with protected attributes (e.g., gender) based on GloVe~\cite{glove}.\footnote{The code and data are released at \url{https://github.com/uclanlp/gn\_glove}} 
GN-GloVe represents protected attributes in certain dimensions while neutralizing the others during  training. As the information of the protected attribute is restricted in  certain dimensions, it can be removed from the embedding easily. By jointly identifying gender-neutral words while learning word vectors, 
GN-GloVe does not require a separate classifier to identify  gender-neutral words; therefore, the error propagation issue is eliminated. 
The proposed approach is generic and can be incorporated with other word embedding models and be
applied in reducing other societal stereotypes.

Our contributions are summarized as follows: 1) To our best knowledge, GN-GloVe is the first method to learn word embeddings with protected attributes;
2) By capturing protected attributes in certain dimensions, our approach ameliorates the interpretability of word representations;
3) Qualitative and quantitative experiments demonstrate that GN-GloVe effectively isolates the protected attributes and preserves the word proximity. 





%% file: related_work.tex
\section{Related Work}
\label{sec:related_work}
\paragraph{Word Embeddings}
Word embeddings serve as a fundamental building block for a broad range of NLP applications~\cite{sentiment,machinetranslation,realtionextraction}
 and various approaches~\cite{mikolov,glove,yoav} have been proposed for training the word vectors. 
Improvements have been made by leveraging semantic lexicons and morphology~\cite{luong2013better,faruqui2014}, disambiguating multiple senses~\cite{vsuster2016bilingual,arora2018linear,upadhyay2017beyond}, and modeling contextualized information by deep neural networks~\cite{elmo}.  
However, none of these works attempts to tackle the problem of stereotypes exhibited in embeddings.

\paragraph{Stereotype Analysis}
Implicit stereotypes have been observed in applications such as online advertising systems~\cite{sweeney2013discrimination}, web search~\cite{kay2015unequal}, and online reviews~\cite{wallace2016jerk}.
Besides, \newcite{jieyu} and \newcite{Rachel18} show that  coreference resolution systems are gender biased. The systems can successfully predict the link between ``the president'' with male pronoun but fail with the female one.  \newcite{rudinger2017social} use point-wise mutual information
to test the SNLI~\cite{snli:emnlp2015} corpus and demonstrate gender stereotypes as well as varying degrees of
racial, religious, and age-based stereotypes in the corpus. A temporal analysis about word embeddings ~\cite{garg2018word} captures changes in gender and ethnic stereotypes over time.
Researchers attributed such problem partly to the biases in the datasets~\cite{jieyu,yao2017beyond} and word embeddings~\cite{garg2017word,caliskan2017semantics} but did not provide constructive solutions.

%% file: methodology.tex
\section{Methodology}
\label{sec:methodology}
In this paper, we take GloVe~\cite{glove} as the base embedding model and gender as the protected attribute. 
It is worth noting that our approach is general and can be applied to other embedding models and attributes. 
Following GloVe~\cite{glove}, we construct a word-to-word co-occurrence matrix $X$, denoting the frequency of the $j$-th word appearing in the context of the $i$-th word as $X_{i,j}$. 
$w, \tilde{w} \in \rnR^{d}$  stand for the embeddings of a center and a context word, respectively, where $d$ is the dimension.

In our embedding model, a word vector $w$ consists of two parts $w=[w^{(a)};w^{(g)}]$.
$w^{(a)}\in\rnR^{d-k}$ and $w^{(g)}\in\rnR^{k}$ stand for neutralized and gendered components respectively,
where $k$ is the number of dimensions reserved for gender information.\footnote{We set $k=1$ in this paper.}
Our proposed gender neutralizing scheme is to reserve the gender feature, known as ``protected attribute'' into $w^{(g)}$. 
Therefore, the information encoded in $w^{(a)}$ is  independent of gender influence. 
We use $v_g \in \rnR^{d-k}$ to denote the direction of gender in the embedding space.
We categorize all the vocabulary words into three subsets: male-definition $\Omega_M$, female-definition $\Omega_F$, and gender-neutral $\Omega_N$,  based on their definition in  WordNet~\cite{miller1998wordnet}.



\paragraph{Gender Neutral Word Embedding}
Our minimization objective is designed in accordance with above insights. It contains three components: \begin{equation}
\label{eq:full_loss}
  J = J_G + \lambda_d J_D +\lambda_e J_E,
\end{equation}
where $\lambda_d$ and $\lambda_e$ are hyper-parameters.


The first component $J_G$ is originated from GloVe~\cite{glove}, which captures the word proximity:
\begin{equation*}
    J_G \!=\! \sum_{i,j=1}^{V}f(X_{i,j})\left(w_i^T\tilde{w}_j + b_i + \tilde{b}_j -\log X_{i,j}\!\right)^2.
    \vspace{-2pt}
\end{equation*}
Here $f(X_{i,j})$ is a weighting function to reduce the influence of extremely large co-occurrence frequencies. $b$ and $\tilde{b}$ are the respective linear biases for $w$ and $\tilde{w}$. 

The other two terms are aimed to restrict gender information in $w^{(g)}$, such that  $w^{(a)}$ is neutral.  Given male- and female-definition seed words $\Omega_M$ and $\Omega_F$, 
we consider two distant metrics and form two types of objective functions.

In $J_D^{L1}$, we directly minimizing the negative distances between words in the two groups: 
$$J_D^{L1} = - \left\|\sum_{w\in \Omega_M} w^{(g)}  - \sum_{w\in \Omega_F} w^{(g)}\right\|_1. $$
In $J_D^{L2}$, we restrict the values of word vectors in $[\beta_1, \beta_2]$ and push $w^{(g)}$ into one of the extremes:
\begin{equation*}
    J_D^{L2} \!=\! \sum_{w\in \Omega_M}\!\!\left\|\beta_1 \bm{e} - w^{(g)}\right\|_2^2 + \sum_{w \in \Omega_F}\!\!\left\| \beta_2\bm{e} - w^{(g)}\right\|_2^2,
    \vspace{-3pt}
\end{equation*}
where $\bm{e} \in \mathcal{R}^{k}$ is a vector of all ones.
 $\beta_1$ and $\beta_2$ can be arbitrary values, and we set them to be $1$ and $-1$, respectively.



Finally, for words in $\Omega_N$, 
the last term encourages their $w^{(a)}$ to be retained in the null space of the gender direction $v_g$:
\begin{equation*}
    J_E = \sum_{w\in \Omega_N}\left(v_g^T w^{(a)}\right)^2,
    \vspace{-3pt}
\end{equation*}
where $v_g$ is estimating on the fly by averaging the differences between female words 
and their male counterparts in a predefined set,
\begin{equation*}
v_g = \frac{1}{|\Omega'|}\sum_{(w_m,w_f)\in \Omega'} (w_m^{(a)} - w_f^{(a)}),
\vspace{-3pt}
\end{equation*}
where $\Omega'$ is a set of predefined gender word pairs.


We use stochastic gradient descent to optimize Eq.~\eqref{eq:full_loss}. To reduce the computational complexity in training the wording embedding, we assume $v_g$ is a fixed vector (i.e., we do not derive gradient w.r.t $v_g$ in updating $w^{(a)}, \forall w\in \Omega'$) and estimate $v_g$ only at the beginning of each epoch.

%% file: experiments.tex
\section{Experiments}
\label{sec:experiments}

In this section, we conduct the following qualitative and quantitative studies: 1) We visualize the embedding space and show that GN-GloVe separates the protected gender attribute from other latent aspects; 2) We measure the ability of GN-GloVe to distinguish between gender-definition words and gender-stereotype words on a newly annotated dataset;  3) We evaluate GN-GloVe on standard word embedding benchmark datasets and show that it performs well in estimating word proximity; 4) We demonstrate that GN-Glove reduces gender bias on a downstream application, coreference resolution.

We compare GN-GloVe with two embedding models, GloVe and Hard-GloVe. GloVe is a widely-used model~\cite{glove}, and we apply the post-processing step introduced in~\cite{kaiwei} to reduce gender bias in GloVe and name it after Hard-GloVe.
All the  embeddings are trained on \textit{2017 English Wikipedia dump} with the default hyper-parameters decribed in ~\cite{glove}.
When training GN-GloVe, we constrain the value of each dimension within $[-1, 1]$ to avoid numerical difficulty.
We set $\lambda_d$ and $\lambda_e$ both to be 0.8. In our preliminary study on development data, we observe that the model is not sensitive to these parameters. 
Unless other stated, we use $J_D^{L1}$ in the GN-GloVe model.

\begin{figure*}[!t]
    \begin{subfigure}[b]{0.32\textwidth}
        \hspace{-0.7em}\includegraphics[width=1.14\textwidth]{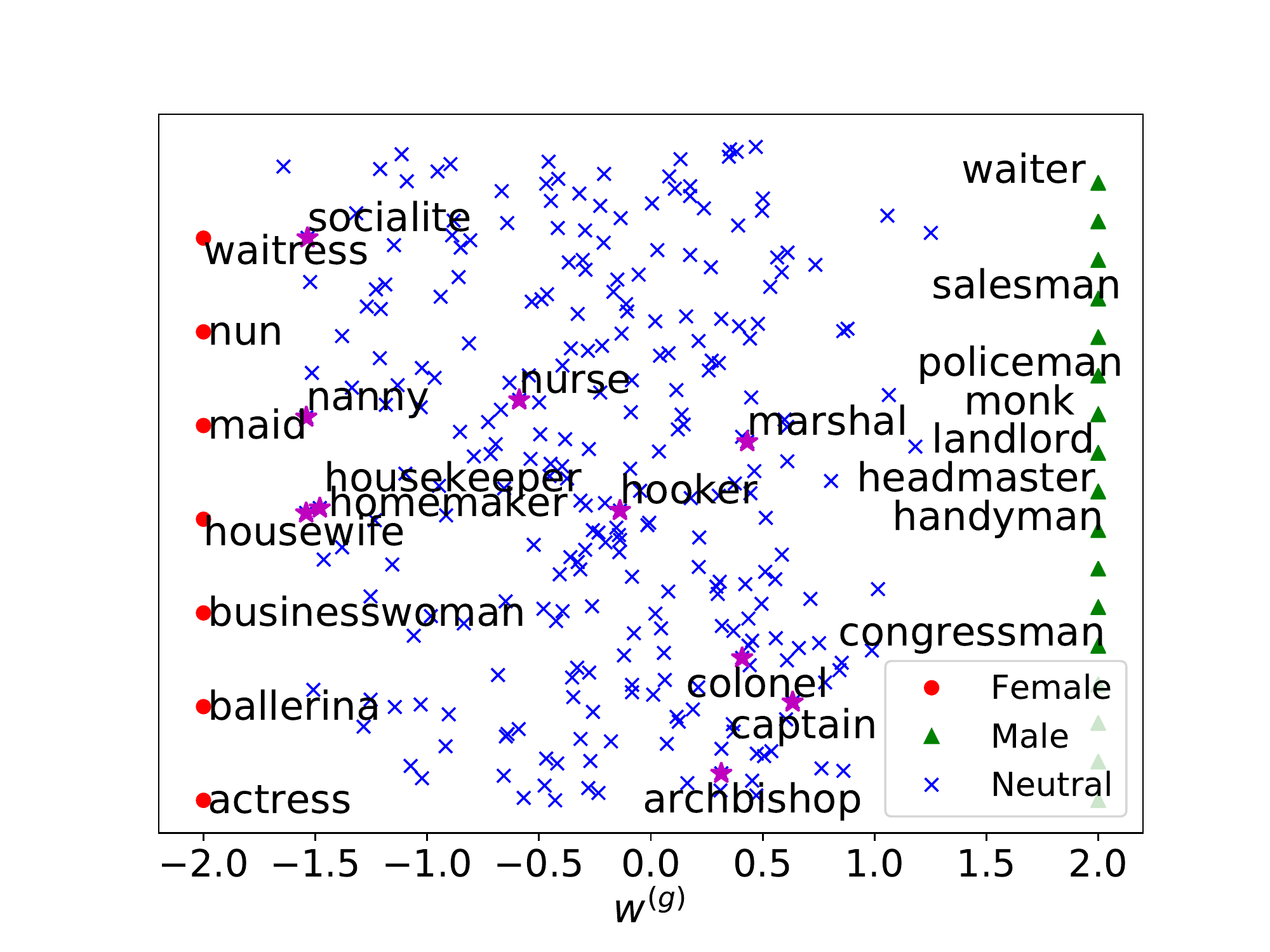}
        \caption{
        $w^{(g)}$ dimension for all the professions
        }
        \label{fig:Wg}
    \end{subfigure}
    ~
    \begin{subfigure}[b]{0.32\textwidth}
        \hspace{-1em}\includegraphics[width=1.14\textwidth]{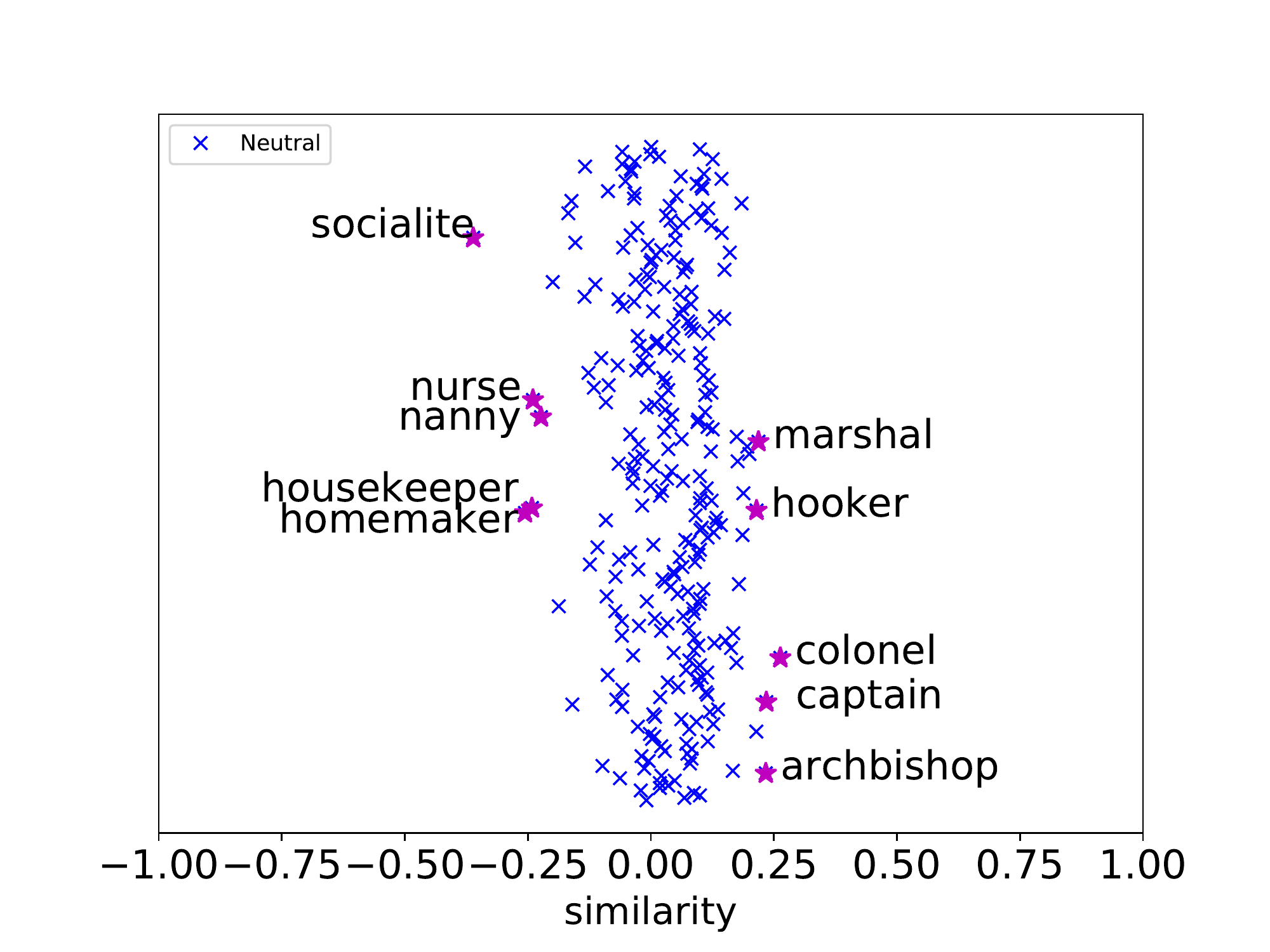}
        \caption{
        Gender-neutral profession words projected to gender direction in GloVe
        }
        \label{fig:GloVe}
    \end{subfigure}
    ~ 
    \begin{subfigure}[b]{0.32\textwidth}
        \hspace{-0.8em}\includegraphics[width=1.14\textwidth]{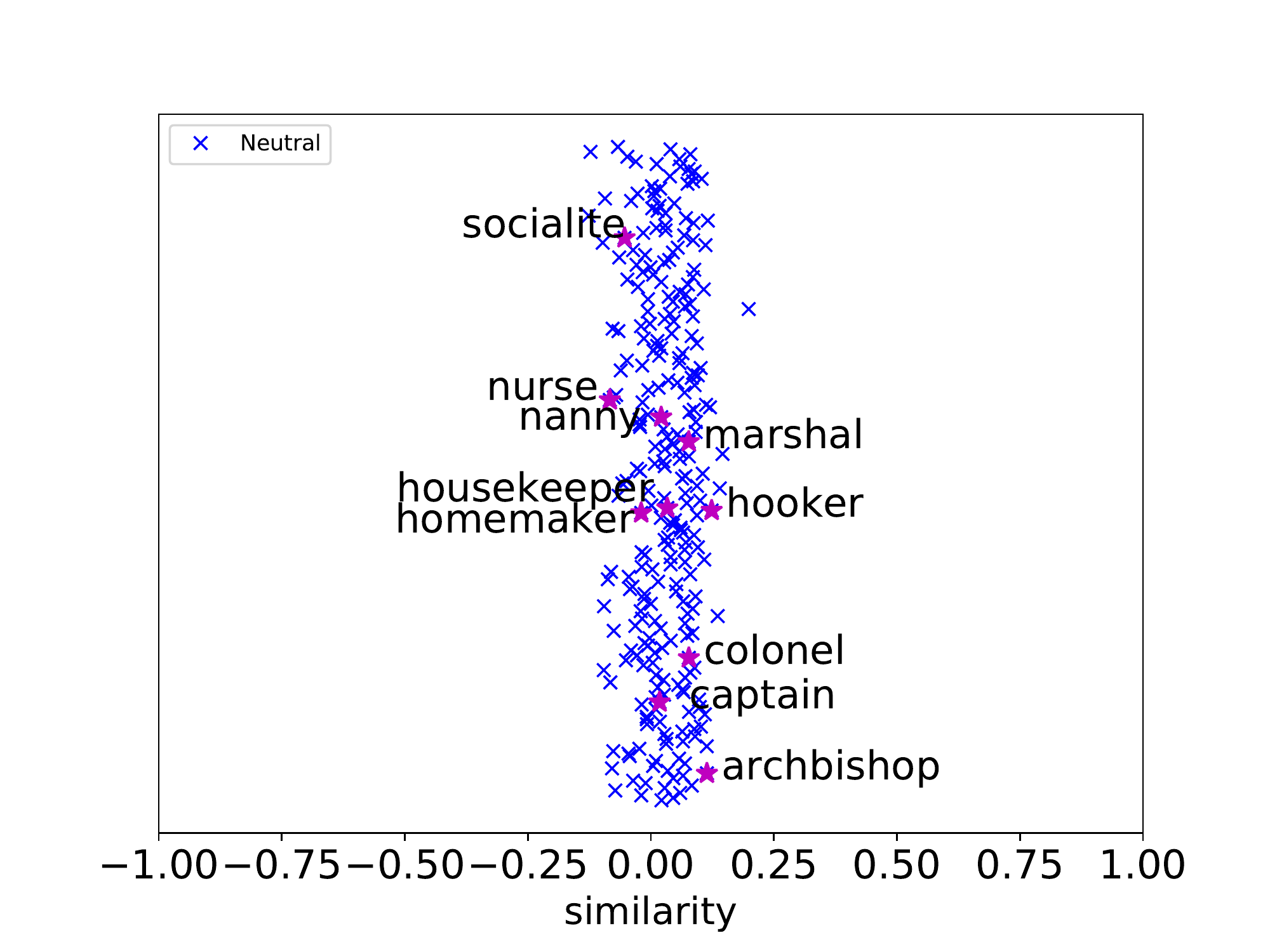}
        \caption{
        Gender-neutral profession words projected to gender direction in  GN-GloVe
        }
        \label{fig:De-GloVe}
    \end{subfigure}
    \caption{
    Cosine similarity between the gender direction and the embeddings of gender-neutral words. 
    In each figure, negative values represent a bias towards female, otherwise male.
    }
    \label{fig:comparison}
\end{figure*}

\paragraph{Separate protected attribute} 
First, we demonstrate that GN-GloVe preserves the gender association (either definitional or stereotypical associations) in $w^{(g)}$\footnote{We follow the original GloVe implementation using the summation of word vector and context vector to represent a word. Therefore, the elements of the word vectors are constrained in [-2, 2]}.
To illustrate the distribution of gender information of different words, we plot 
Fig.~\ref{fig:Wg} using $w^{(g)}$ for the x-axis and a random value for the y-axis to spread out words in the plot. As shown in the figure, the gender-definition words, e.g. ``waiter'' and ``waitress'', fall far away from each other in $w^{(g)}$. In addition, words such as ``housekeeper'' and ``doctor'' are inclined to different genders and their $w^{(g)}$ preserves such information. 

Next, we demonstrate that GN-GloVe reduces gender stereotype using a list of profession titles from~\cite{kaiwei}. All these profession titles are neutral to gender by definition.
In Fig.~\ref{fig:GloVe} and Fig.~\ref{fig:De-GloVe},
we plot the cosine similarity between each word vector $w^{(a)}$ and the gender direction ${v_g}$ (i.e.,  $\frac{w^T v_g}{\norm{w}\norm{v_g}}$). Result shows that words, such as ``doctor'' and ``nurse'', possess no gender association by definition, but their GloVe word vectors exhibit strong gender stereotype. In contrast, the gender projects of GN-GloVe word vectors $w^{(a)}$ are closer to zero. This demonstrates the gender information has been substantially diminished from $w^{(a)}$ in the GN-GloVe embedding.


We further quantify the gender information exhibited in the embedding models.
For each  model, we project the word vectors of occupational words into the gender sub-space defined by ``he-she'' and compute their average size.   A larger projection indicates an embedding model is more biased. Results show that the average projection of GloVe is 0.080, the projection of Hard-GloVe is 0.019, and the projection of Gn-Glove is 0.052. Comparing with GloVe, GN-GloVe reduces the bias by $35\%$. 
Although Hard-GloVe contains less gender information, we will show later GN-GloVe can tell difference between gender-stereotype and gender-definition words better.

\paragraph{Gender Relational Analogy} 
To study the quality of the gender information present in each model, we follow SemEval 2012 Task2~\cite{jurgens2012semeval} to create an analogy dataset, \textit{SemBias}, with the goal to identify the correct analogy of ``he - she'' from four pairs of words. Each instance in the dataset consists of four word pairs: a gender-definition word pair (Definition; e.g., ``waiter - waitress''), a gender-stereotype word pair (Stereotyp; e.g., ``doctor - nurse'') and two other pairs of words that have similar meanings (None; e.g.,  ``dog - cat'', ``cup - lid'')\footnote{The pair is sampled from the list of word pairs with ``SIMILAR: Coordinates'' relation annotated in ~\cite{jurgens2012semeval}. The original list has 38 pairs. After removing gender-definition word pairs, 29 are left.}. We consider 20 gender-stereotype word pairs and 22 gender-definition word pairs and use their Cartesian product to generate 440 instances. Among the 22 gender-definition word pairs, there are 2 word pairs that are not used as a seed word during the training. To test the generalization ability of the model, we generate a subset of data  (SemBias (subset)) of 40 instances associated with these 2 pairs.

Table~\ref{tab:isSim} lists the percentage of times that each class of pair is on the top based on a word embedding model~\cite{mikolov2013linguistic}. GN-GloVe achieves 97.7\% accuracy in identifying gender-definition word pairs as an analogy to ``he - she''. In contrast, GloVe and Hard-GloVe makes significantly more mistakes. On the subset, GN-GloVe also achieves significantly better performance than Hard-Glove and GloVe, indicating that it can generalize the gender pairs on the training set to identify other gender-definition word pairs.

\begin{table}
\centering
\resizebox{\linewidth}{!}{
\begin{tabular}{ll|ccc}
\hline
Dataset & Embeddings & Definition & Stereotype & None\\
\hline
\multirow{ 3}{*}{SemBias} & GloVe      & 80.2       & 10.9       & 8.9 \\
& Hard-Glove &  84.1       &   6.4       &  9.5 \\
& GN-GloVe   &   97.7      &    1.4        &  0.9  \\
\hline
\multirow{ 3}{*}{\shortstack[l]{SemBias\\(subset)}} & GloVe      &  57.5       &   20      & 22.5 \\
& Hard-Glove &    25     &  27.5       & 47.5 \\
& GN-GloVe   & 75        & 15          & 10  \\
\hline
\end{tabular}
}
\caption{Percentage of predictions for each category on gender relational analogy task. 
}
\label{tab:isSim}
\end{table}

\begin{table*}[!thbp]
\centering
\resizebox{\linewidth}{!}{
\begin{tabular}{l|cc|cccccc}
\hline
\multirow{2}{*}{Embeddings} & \multicolumn{2}{c|}{Analogy} & \multicolumn{6}{c}{Similarity} \\
  & Google & MSR & WS353-ALL  & RG-65  & MTurk-287 & MTurk-771 & RW & MEN-TR-3k \\
\hline
GloVe      &    \textbf{70.8} & \textbf{45.8 }    & 62.0    & 75.3   & 64.8    & 64.9   & 37.3    & 72.2 \\
Hard-GloVe &   \textbf{70.8} & \textbf{45.8}     & 61.2    & 74.8   & 64.4    & 64.8   & 37.3    & 72.2 \\
GN-GloVe-L1&   68.9 & 43.7   & \textbf{62.8}   & 74.1   & 66.2    & \textbf{66.2}   & \textbf{40.0}    & \textbf{74.5} \\
GN-GloVe-L2&   68.8 & 43.6   & 62.5  & \textbf{76.4}   & \textbf{66.8}    & 65.6   & 39.3    & 74.4 \\

\hline
\end{tabular}
}
\caption{Results on the benchmark datasets. Performance is measured in accuracy
and in Spearman rank correlation for word analogy and word similarity tasks, respectively.}
\label{tab:sims2}
\end{table*}

\paragraph{Word Similarity and Analogy} In addition, we evaluate the word embeddings on the benchmark tasks to ensure their quality. 
The word similarity tasks measure how well a word embedding model captures the similarity between words comparing to human annotated rating scores. Embeddings are tested on multiple datasets:  WS353-ALL~\cite{finkelstein2001placing}, RG-65~\cite{rubenstein1965contextual},  MTurk-287~\cite{radinsky2011word}, MTurk-771~\cite{halawi2012large}, RW~\cite{luong2013better}, and MEN-TR-3k~\cite{bruni2012distributional} datasets.  
The analogy tasks are to answer the question ``\textit{A} is to \textit{B} as \textit{C} is to \underline{\hbox to 2mm{}}?'' by finding a word vector $w$ that is closest to $w_A - w_B + w_C$ in the embedding space.
Google~\cite{mikolov2013efficient} and MSR~\cite{mikolov2013linguistic} datasets are utilized for this evaluation. 
The results are shown in Table ~\ref{tab:sims2}, where the suffix ``-L1'' and ``-L2'' of GN-GloVe stand for the GN-GloVe  using  $J_D^{L1}$ and $J_D^{L2}$, respectively. 
Compared with others, GN-GloVe achieves a higher accuracy in the similarity tasks and its analogy score slightly drops indicating that GN-GloVe is capable of preserving proximity among words. 

\paragraph{Coreference Resolution} Finally, we investigate how the gender bias in word embeddings affects a downstream application, such as coreference resolution.
Coreference resolution aims at clustering the denotative noun phrases referring to the same entity in the given text. We evaluate our models on the Ontonotes 5.0~\cite{weischedel2012ontonotes} benchmark dataset and the WinoBias dataset~\cite{zhao2018gender}.\footnote{Specifically, we conduct experiments on the Type 1 version. } In particular, the WinoBias dataset is composed of pro-stereotype (PRO) and anti-stereotype (ANTI) subsets. 
 The PRO subset consists of sentences where a gender pronoun refers to a profession, which is dominated by the same gender.
 Example sentences include ``The CEO raised the salary of the receptionist because he is generous.'' In this sentence, the pronoun ``he'' refers to ``CEO'' and this reference is consistent with societal stereotype. The ANTI subset contains the same set of sentences, but the gender pronoun in each sentence is replaced by the opposite gender. For instance, the gender pronoun ``he'' is replaced by ``she'' in the aforementioned example. Despite the sentence is almost identical, the gender pronoun now refers to a profession that is less represented by the gender.  Details about the dataset are in ~\cite{zhao2018gender}.

We train the end-to-end coreference resolution model~\cite{lee2017end} with different word embeddings on OntoNote and report their performance in Table \ref{tab:coref}.
For the WinoBias dataset, we also report the average (Avg) and absolute difference (Diff) of F1 scores on two subsets. A smaller Diff value indicates less bias in a system.
Results show that GN-GloVe achieves comparable performance as Glove and Hard-GloVe on the OntoNotes dataset while distinctly reducing the bias on the WinoBias dataset. 
When only the $w^{(a)}$ potion of the embedding is used in representing words, GN-GloVe($w^{(a)}$) further reduces the bias in coreference resolution. 


\begin{table}
\centering
\resizebox{\linewidth}{!}{
\begin{tabular}{l|c|cccc} \hline
Embeddings     &  OntoNotes-test & PRO & ANTI & Avg & Diff\\ 
\hline
GloVe          & 66.5       &  76.2 & 46.0 & 61.1 & 30.2\\
Hard-Glove     & 66.2       & 70.6 & 54.9 & 62.8 & 15.7\\
GN-GloVe       & 66.2       & 72.4 & 51.9  & 62.2 & 20.5 \\
GN-GloVe($w_a$) & 65.9       & 70.0 & 53.9 & 62.0 & 16.1\\
\hline
\end{tabular}
}
\caption{F1 score (\%) on the coreference system. 
}
\vspace{-10pt}
\label{tab:coref}
\end{table}


%% file: conclusion.tex
\section{Conclusion and Discussion}
\label{sec:conclusion}
In this paper, we introduced an algorithm for training gender-neutral word embedding. Our method is general and can be applied in any language as long as a list of gender definitional words is provided as seed words (e.g., gender pronouns). Future directions include extending the proposed approach to model other properties of words such as sentiment and generalizing our analysis beyond binary gender. 

\section*{Acknowledgement}
This work was supported by National Science Foundation Grant IIS-1760523. We would like to thank James Zou, Adam Kalai, Tolga Bolukbasi, and Venkatesh Saligrama for helpful discussion, and anonymous reviewers for their feedback.
